\documentclass{article}

\usepackage{microtype}
\usepackage{graphicx}
\usepackage{booktabs} 
\usepackage{subcaption}
\usepackage{lscape}

\usepackage{amsmath}
\usepackage{amssymb}
\usepackage{bm}

\usepackage{xr}

\newcommand\leftidx[3]{%
  {\vphantom{#2}}#1#2#3%
}

\newtheorem{proposition}{Proposition}
\newtheorem{definition}{Definition}
\newtheorem{lemma}{Lemma}

\newcommand{\ImportanceLandscape}{importance landscape }
\newcommand{\FeatureImportance}{unit importance }

\def\eqref#1{equation~\ref{#1}}

\def\1{\bm{1}}
\def\0{\bm{0}}

\DeclareMathAlphabet{\mathsfit}{\encodingdefault}{\sfdefault}{m}{sl}
\SetMathAlphabet{\mathsfit}{bold}{\encodingdefault}{\sfdefault}{bx}{n}
\newcommand{\tens}[1]{\bm{\mathsfit{#1}}}

\newcommand{\softmax}{\mathrm{softmax}}

\newcommand{\qedwhite}{\hfill \ensuremath{\Box}}

\DeclareMathOperator*{\argmax}{arg\,max}
\DeclareMathOperator*{\argmin}{arg\,min}

\newcommand{\path}[2]{\ensuremath{\overline{p}_{#1 \to #2}}}

\newcommand{\norm}[1]{\left\lVert#1\right\rVert}

\newcommand{\comments}[1]{}

\usepackage{hyperref} 

\usepackage[accepted]{icml2020_edited}

\makeatletter
\newcommand*{\addFileDependency}[1]{
  \typeout{(#1)}
  \@addtofilelist{#1}
  \IfFileExists{#1}{}{\typeout{No file #1.}}
}
\makeatother

\icmltitlerunning{\P ILCRO}

\begin{document}

\twocolumn[
\icmltitle{\P ILCRO: Making Importance Landscapes Flat Again}



\icmlsetsymbol{equal}{*}

\begin{icmlauthorlist}
\icmlauthor{Vincent Moens}{aig}
\icmlauthor{Simiao Yu}{aig}
\icmlauthor{Gholamreza Salimi-Khorshidi}{aig}
\end{icmlauthorlist}

\icmlaffiliation{aig}{Investments AI, AIG, London, United Kingdom}

\icmlcorrespondingauthor{Vincent Moens}{vincent.moens@gmail.com}

\icmlkeywords{Computer Vision, Convolutional Neural Networks, Regularisation}

\vskip 0.3in
]



\printAffiliationsAndNotice{}  

\begin{abstract}
Convolutional neural networks have had a great success in numerous tasks, including image classification, object detection, sequence modelling, and many more.
It is generally assumed that such neural networks are translation invariant.
This paper shows that most of the existing convolutional architectures define, at initialisation, a specific feature importance landscape that conditions their capacity to attend to different locations of the images later during training or even at test time.
We demonstrate how this phenomenon occurs under specific conditions and how it can be adjusted under some assumptions.
We derive the \P-objective, or PILCRO,
a simple regularisation technique that favours weight configurations that produce smooth, low-curvature importance landscapes that are conditioned on the data and not on the chosen architecture.
Through extensive experiments, we further show that \P-regularised versions of popular computer vision networks have a flat importance landscape, train faster, result in a better accuracy and are more robust to noise at test time, when compared with their original counterparts in common computer-vision classification settings.
\end{abstract}

\section{Introduction}

Following birth, the human infant peripheral vision is surprisingly weak: while the morphological maturity of the peripheral retinal structures is more advanced than the foveal, the corresponding functional maturity is poorer \cite{Courage1996}.
Although this difference is partially mitigated after a few years, the fovea keeps playing a prominent role in the human vision system, while the peripheral cells of the retina are mostly involved in secondary tasks, such as movement perception and night vision.

In this paper, we show that a similar development occurs during training of convolutional neural networks: when initialized, most networks concentrate their attention on the center of the images and on other locations defined by the network topology, and later during training, learn to attend to peripheral features when necessary.


More specifically, Convolutional Neural Networks (CNNs) usually have designs that are inherently biased towards paying more attention to subsets of the pixels which are typically located in, but not limited to, the center of the images.
Whereas our poor peripheral visual accuracy is compensated by the capacity of the human eye to actively explore the visual field, CNNs are trained on a set of fixed images, and lack the capability to orientate their attention to the relevant parts of their visual field.

Also, aside from trivial tasks such as classifying black-and-white digits on a unicolor background, the relevant signal (the feature) is usually accompanied by other non-relevant signals, e.g. clouds next to a plane and branches and leaves next to a bird. If the relevant signal does not match the preferred location of the network, there is a risk that confounding signals will prevail, and hence generalisation will be harmed.

The propensity of CNNs to consider some pixel locations as more salient than others can be expected to have a series of negative impacts on training quality:
\begin{itemize}
    \item \textbf{Bias to some feature location}: By paying more attention to some parts of the image, CNNs implicitly assume that features are more likely to occur in specific areas than in others. Yet, this distribution is dictated by the architecture and not by the data.
    \item \textbf{Overfitting and noise robustness}: It is likely that smooth, flat or data-dictated feature importance landscapes will be less sensitive to the location of the feature of interest in the training dataset.
    Similarly, noisy input images should be better classified if the class is predicted based on a larger and more optimal cluster of pixels, looking at the entirety of the image rather than a predefined subset of its pixels.
    \item \textbf{Training speed}: A model that is biased to look at some parts of the image may be slower to train if a subset of images violates its default salience map, as it will need to overcome this problem as learning goes.
    
\end{itemize}

Our contributions to the field are the following: first, we formally show that most conventional CNNs will usually consider some input units (i.e. pixels) as more salient than others in a way that is conditioned on the architecture being used and on the size of the image.
Next, we introduce the PILCRO-objective (or \P-objective), a simple regularisation loss that comes in two flavours: the \P$_c$-objective maximises the smoothness of the importance landscape whereas its counterpart \P$_t$-objective minimises the distance between the importance landscape and a location-based smoothness measure of the training dataset, thereby indirectly minimizing the curvature of the importance landscape in a data-driven manner.

The paper is structured as follows. In Section~\ref{sec:methods}, we will approach the problem theoretically using a simplified mathematical setting that will be made more complex later on.
Next, in Section~\ref{sec:solution}, we will present a simple family of regularisation losses designed to flatten the \ImportanceLandscape of the input images.
Section~\ref{sec:results} will then show results of our regularisation methods on various classification tasks in terms of accuracy and noise robustness.

\section{Related work}\label{sec:relatedwork}
It has been shown that, under specific conditions, CNNs have the desirable property of being translation invariant \cite{Wiatowski2018}. However, the proof of such assertion assumes that there is a clear distinction between the signal to classify and the background (for instance, in the case of MNIST \cite{lecun-mnisthandwrittendigit-2010} dataset).
In real settings, however, multiple signals coexist, and translating a feature usually involves changing the content of the whole image.
The ability of CNNs to accurately classify a translated images in such settings is therefore less clear.

Our work is closely related to the assessment of feature importance, a hot topic in explainable Artificial Intelligence, where the general hope is to retrieve from a trained network the set of pixels that were used for inference. 
\cite{Breiman2001} suggested that randomly shuffling units (e.g. pixels in the present setting) across an input location might be used to assess this location salience; however, this technique is disputed by recent works \cite{Hooker2019}. 
Other methods \cite{Lundberg2017} rely on feature hold-on estimation or require expensive post-training assessment \cite{Ribeiro2016}, and are hence difficult to combine with training.
Some techniques \cite{Lu2018} integrate the feature importance measuring tool in the network, thereby limiting these measures to a specific subset of models.
Techniques involving gradient intensity \cite{Baehrens2010,Smilkov2017,Adebayo2018,Sundararajan2017} have gained popularity recently due to their straightforward usage.
Our work subscribes to this logic, to the extent that we use gradient intensities to measure the quantity of information flowing from the a pixel to the last prediction unit of a CNN. However, it differs in the sense that we are not interested in the image-wise feature importance heatmap, but rather how a similar heatmap would behave on average over a whole dataset. 
It has been shown that, under specific conditions, CNNs have the desirable property of being translation invariant \cite{Wiatowski2018}. However, the proof of such assertion assumes that there is a clear distinction between the signal to classify and the background (for instance, in the case of MNIST \cite{lecun-mnisthandwrittendigit-2010} dataset).
In real settings, however, multiple signals coexist, and translating a feature usually involves changing the content of the whole image.
The ability of CNNs to accurately classify a translated images in such settings is therefore less clear.

Our work is closely related to the assessment of feature importance, a hot topic in explainable Artificial Intelligence, where the general hope is to retrieve from a trained network the set of pixels that were used for inference. 
\cite{Breiman2001} suggested that randomly shuffling units (e.g. pixels in the present setting) across an input location might be used to assess this location salience; however, this technique is disputed by recent works \cite{Hooker2019}. 
Other methods \cite{Lundberg2017} rely on feature hold-on estimation or require post-training expensive assessment \cite{Ribeiro2016}, and are hence difficult to combine with training.
Some techniques \cite{Lu2018} integrate the feature importance measuring tool in the network, thereby limiting these measures to a specific subset of models.
Techniques involving gradient intensity \cite{Baehrens2010,Smilkov2017,Adebayo2018,Sundararajan2017} have gained popularity recently due to their straightforward usage.
Our work subscribes to this logic, to the extend that we use gradient intensities to measure the quantity of information flowing from the a pixel to the last prediction unit of a CNN. However, it differs in the sense that we are not interested in the image-wise feature importance heatmap, but rather how a similar heatmap would behave on average over a whole dataset. 

\comments{--Depending on the results, add something about Adversarial attacks on CNNs--}

\section{Methods}\label{sec:methods}

\subsection{The number of paths in convolutional networks}
\begin{figure}
  \centering
  \captionsetup{width=0.8\linewidth}
  \includegraphics[width=0.9\linewidth]{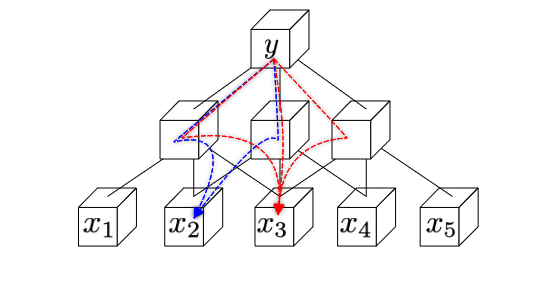}
  \caption{Number of paths in a simple 2-layer 1d-CNN with kernel of size 3. The center pixel has three paths linking it to the output unit (red), whereas the pixel next to it has only two (blue).}
  \label{fig:fig1a}
\end{figure}

\begin{figure}
\centering
  \captionsetup{width=0.8\linewidth}
  \includegraphics[width=0.9\linewidth]{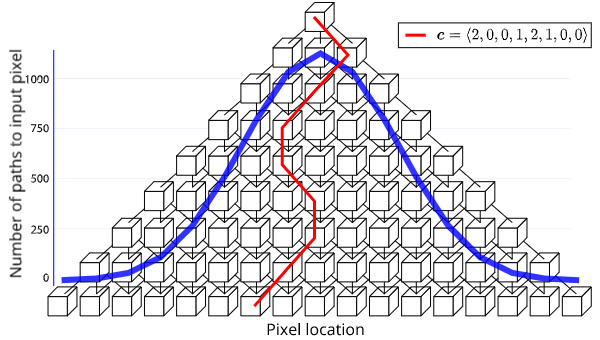}
  \caption{According to the central limit theorem, if the network is regular, as its depth grows, the number of paths from the center output unit to the input units approaches the shape of a Gaussian distribution centered on the central input unit. The red path shows an example of a command vector such that $route_{y \rightarrow \textbf{x}}(\textbf{c}) = 6$.}
  \label{fig:fig1b}
\end{figure}

CNNs define a backward forking paths topology (which we will refer to as a \textit{lattice}) whose number of paths to the root units, in its simplest forms (e.g. Figure~\ref{fig:fig1b}), behaves like a Generalized Pascal triangle.
To show this, take the archetypal 1-dimensional CNN 
displayed in Figure~\ref{fig:fig1a}.
One can easily see that the number of paths from the output tensor to a single unit of the input tensor is greater for the input units located at the center than for units situated on the edges.

With this simple structure in place, we make explicit the notation that will be used before proceeding to a rigorous analysis of the number of paths connecting two nodes in this lattice.
For simplicity, this section assumes one-dimensional CNNs (unless specified otherwise) of depth $L$, input tensor $\tens{X} \in \mathbb{R}^{d_x \times C_{\text{in}} \times B}$ and output $\tens{H}^{(L)} \in \mathbb{R}^{d_y \times C_{\text{out}} \times B}$, where $B$ is the batch size, $d_x$ and $d_y$ are the dimension of the input and output, respectively.
Also, intermediate cell values are indicated by $\tens{H}^{(l)}$, where $l \in \{l\}_{l=0}^L$ is the index of the layer in the network (hence $\tens{X} \equiv \tens{H}^{(0)}$).
Lower indices indicate the unit index of the tensor, that is $\tens{H}^{(0)} \equiv \left\langle\bm{H}^{(0)}_0, \bm{H}^{(0)}_1, \dots, \bm{H}^{(0)}_{d_x-1}\right\rangle$.

In its simplest form, we define a size $K$, $C_{\text{in}}$ to $C_{\text{out}}$ one-dimensional convolutional layer of weight $\tens{W}^{(l)} \in \mathbb{R}^{K \times C_{\text{out}}  \times C_{\text{in}}}$ and bias $\bm{b}^{(l)} \in \mathbb{R}^{1 \times C_{\text{out}}}$ on an input tensor $\tens{H}^{(l)} \in \mathbb{R}^{N^{(l)} \times C_{\text{in}} \times B}$ as the following operation:
\begin{equation*}
\begin{aligned}
    \tens{A}^{(l+1)} &=\tens{W}^{(l)} * \tens{H}^{(l)} + \bm{b}^{(l)}\\
    &= \left\langle \bm{A}^{(l+1)}_0, \bm{A}^{(l+1)}_1, \dots, \bm{A}^{(l+1)}_{N-k} \right\rangle
\end{aligned}
\end{equation*}
where
\begin{equation*}
\begin{aligned}
    \bm{A}^{(l+1)}_m 
    &=\sum_{k=0}^{K-1} \bm{W}_{k}^{(l)} \bm{H}_{m+k}^{(l)} + \bm{b}^{(l)}
\end{aligned}
\end{equation*}
If nonlinearities are present, then $\tens{H}^{(l)} = \sigma\left(\tens{A}^{(l)}\right)$, otherwise $\tens{H}^{(l)} = \tens{A}^{(l)}$. 
For simplicity, throughout the rest of the paper, we will assume a dilation and a stride of one, unless specified otherwise.

Next, for the purpose of measuring the importance of any given input unit, we focus on the problem of routing output to input unit in the network.
For this purpose, we introduce a $route$ function, that given a cell and a preceding layer couple, returns the cell index in this layer for an input vector of commands that reflect the direction taken when going through the network topology with this sequence of commands.
We will then show that the number of unique command vectors that link two cells -- say $a$ and $b$ -- in a network (i.e. that return the $b$ cell index when the $a$ cell is queried together with the $b$ cell layer) impacts how much the $b$ cell influences the value of the $a$ cell at initialization.

\begin{definition}
Given a 1d-CNN composed solely of stacked convolutional layers, the recursive function $route_{h^{(l)} \to \tens{H}^{(l')}}( \bm{C} ) = \mathbb{N}^{L\times 3} \mapsto \mathbb{N}^3$ (which is defined only for $l > l'$) is defined as:
\begin{equation}\label{eq:route}
\begin{aligned}
    &route_{h^{(l)}_{\bm{j}} \to \tens{H}^{(l')}}( \bm{C} ) =\\
    &\qquad \left \{ \begin{array}{ll}
        route_{h_{\bm{j}}^{(l-1)} \to \tens{H}^{(l')}}( \bm{C}^{0:l-1} ), &\\
        \,\bm{j} = route_{h^{(l)}_{\bm{j}} \to \tens{H}^{(l-1)}}( \bm{c}^{(l-l'-1)} ) & \text{ if } l > l'+1\\
        \bm{i}\quad\text{s.t. }h_{\bm{i}}^{(l')} = \frac{\partial h_{\bm{j}}^{(l)}}{\partial w^{(l')}_{\bm{c}^{0}}} & \text{ if } l = l' + 1.
    \end{array} \right.
\end{aligned}
\end{equation}
\end{definition}
In words, the function $route$ takes
as input a vector of ``commands'' $\bm{C}~=~\left\langle \bm{c}^{(0)}, \bm{c}^{(1)}, \dots, \bm{c}^{(l-l'-1)}~\right \rangle
$, each representing an entry index of the corresponding weight tensor. It outputs the index of the entry in the input tensor (at the desired layer $l'$) that equals the partial derivative of the output w.r.t. the weights indexed by their corresponding layer-command.
This index corresponds to the final unit reached when taking the desired route in the network lattice, starting from a given output entry $h^{(l)}_{\bm{j}}$.
An example of a $route$ command vector is displayed in Figures~\ref{fig:fig1a}~and~\ref{fig:fig1b}.

For a given output unit, multiple command vectors may lead to the same input unit: it is straightforward for instance to show that the path in a regular lattice is permutation invariant w.r.t. $\bm{C}$:
\begin{align*}
&route\left(\left \langle \dots, \bm{c}^{(i)}, \bm{c}^{(j)} \dots \right \rangle\right) =\\ &\quad route\left(\left \langle \dots, \bm{c}^{(j)}, \bm{c}^{(i)}, \dots \right \rangle\right) \, \forall i,j \in \{0,1,\dots,L-1\}.
\end{align*}


If needed, this scheme can be slightly altered to account for residual connections
\footnote{We can assume that the set of commands from which $\bm{C}$ is drawn is enriched by a $skipped$ command: $c^{(l)}_0 \in \{0,1,\dots,k-1, skip\}$. 
Similarly, the $route$ operation in Equation~\ref{eq:route} can be adapted to allow for such layer-skipping behaviour when needed.}.

\begin{definition}
We say there is a \textit{path} from $h_{\bm{j}}^{(l)}$ to $h_{\bm{i}}^{(l')}$ in the network lattice \textit{iff.} there exists a vector of commands $\bm{C}$ whose respective $route$ mapping returns $\bm{i}$ when executed across the network:
\begin{equation*}
\exists\,p_{h_{\bm{j}}^{(l)} \to h_{\bm{i}}^{(l')}} \iff \exists\,\bm{C} \in \mathbb{N}^{L\times 3}\,\text{s.t.}\, route_{h_{\bm{j}}^{(l)} \to \tens{H}^{(l')}}( \bm{C} ) = \bm{i}.
\end{equation*}
\end{definition}

To count the number of paths linking two given nodes in a regular lattice, we further define the set $\mathbb{P}_{h_{\bm{j}}^{(l)} \to h_{\bm{i}}^{(l')}}$ of the permutations of single commands that lead to $h_{\bm{i}}^{(l')}$ starting from $h_{\bm{j}}^{(l)}$:
\begin{equation}\label{eq:perm}
\begin{split}
\mathbb{P}_{h_{\bm{j}}^{(l)} \to h_{\bm{i}}^{(l')}} = \bigl\{ \bm{C} \, \bigm| \, route_{h_{\bm{j}}^{(l)} \to \tens{H}^{l'}}( \bm{C} ) = \bm{i},\\ 
 \, \bm{C} \in P(L,K C_{\text{in/out}}^2)  \bigr \}
\end{split}
\end{equation}
where $P(L,K C_{\text{in/out}}^2)$ is the set of $L$ length permutations of $K C_{\text{in/out}}$ items, and we assume that $C_{\text{in/out}}\triangleq C_{\text{in}} = C_{\text{out}}$.
In more complex lattices, $\mathbb{P}_{h_{\bm{j}}^{(l)} \to h_{\bm{i}}^{(l')}}$ can be defined as the set of commands that return $\bm{i}$ when executed across the lattice.
The number of paths linking two nodes is simply defined as the cardinality of this set:
\begin{align*}
    \# p_{h_{\bm{j}}^{(l)} \to h_{\bm{i}}^{(l')}} &\triangleq \left | \mathbb{P}_{h_{\bm{j}}^{(l)} \to h_{\bm{i}}^{(l')}} \right |.
\end{align*}
Usually, for two adjacent layers, there is  at most one path linking two unit entries. 
Yet, an unit can be more or less connected to the adjacent layer. 
We further define the expected number of paths of a layer to an entry as
\begin{align*}
    \# \overline{p}_{\tens{H}^{(L)}\rightarrow h_{\bm{i}}^{l'}} &\triangleq \mathbb{E}_{\bm{j}}\left[\# p_{h_{\bm{j}}^{(L)}\rightarrow h_{\bm{i}}^{l'}}\right]\\&= \frac{1}{Z} \sum_{\bm{j}} \# p_{h_{\bm{j}}^{(L)}\rightarrow h_{\bm{i}}^{l'}},
\end{align*}
which measures the connectivity of a layer to a given preceding entry of layer $l'$ indexed by $\bm{i}$.

Similarly, we define $\mathbb{C}_{h_{\bm{j}}^{(l)} \to h_{\bm{i}}^{(l')}}$ as the set of single command combinations leading to $h_{\bm{i}}^{(l')}$ when starting from $h_{\bm{j}}^{(l)}$:
\begin{equation*}
\begin{split}
\mathbb{C}_{h_{\bm{j}}^{(l)} \to h_{\bm{i}}^{(l')}} =\Bigl\{ \bm{C} \, \bigm| \, route_{h_{\bm{j}}^{(l)} \to \tens{H}^{l'}}( \bm{C} ) = \bm{i},  \\ \, \bm{C} \in C(L,K C_{\text{in/out}})  \bigr\}.
\end{split}
\end{equation*}
We can then count the number of paths linking two given nodes in a regular lattice by using the multinomial coefficient formula:
\begin{align*}  
\# p_{h_{\bm{j}}^{(l)} \to h_{\bm{i}}^{(l')}}&=\sum_{ \bm{c} \in \mathbb{C}_{h_{\bm{j}}^{(l)} \to h_{\bm{i}}^{(l')}}} \frac{L!}{ \prod_{ \bm{j} } \#\left(\bm{C}\right)_{\bm{c}=\bm{j}} !  }
\end{align*}
if the lattice is regular.
This fact has interesting implications: using the central limit theorem, it is not difficult to see that the number of paths from top to bottom of a lattice such as the one displayed in Figure~\ref{fig:fig1b} will follow a shape dictated by a Gaussian bell centered on the middle input units.

\subsection{Input unit relevance in Convolutional Neural Networks}
We now turn to the problem of estimating the importance of an input unit (e.g. a pixel or a voxel) on a model predictions.

Input unit importance (or relevance) can be defined and measured in many different ways.
Intuitively, we could say that a unit is important if modifying it (by adding noise, changing its value, or discarding it) greatly impacts the predictions.
A standard method to assess \FeatureImportance is to look at the absolute change in some measure of the output quality (for instance, prediction accuracy) when it is randomly resampled in the dataset \cite{Breiman2001}, although other \FeatureImportance measures could be considered (see Section~\ref{sec:relatedwork}) but are beyond the scope of this paper.

Under some settings, output-to-input gradients can be used for feature importance assessment \cite{Baehrens2010}.
Interestingly, the following lemma shows that expected square gradient can in some cases approximate the resampling-based unit importance measure propsed above:
\begin{lemma}\label{lemma:1}
Under specific assumptions, including twice differentiability of the network operations, and that the second derivative of this model is zero almost everywhere, a close proxy to the shuffled input importance measure can be obtained via the square root of the expected squared partial derivative of the output w.r.t. the input entry of interest, scaled by its standard deviation.
\end{lemma}
The proof of Lemma~\ref{lemma:1} is given in Appendix~A.

We assume the standard deviation of the input units to be approximately identical across input locations and can be therefore ignored.

The pixel importance is a simple measure of ``how much'' gradient flows from one pixel location to the last layer of the network, which we can expect to loosely correlate with the impact that location will have, on average, over the loss being optimized in a gradient descent setting.
We now turn to the problem of estimating the expected pixel importance over multiple initialisations of a given CNN. 
\begin{proposition}\label{propo:restricted}
In a network of stacked convolutional layers, and under mild assumption about network initialisation, the expected squared partial derivatives of the last layer with respect to an input entry, is linearly proportional to the sum over the set of output-to-input paths of the product of the variances of the weights of the layers crossed through by the respective path:
\begin{equation}\label{eq:prop}
    \mathbb{E}_{\bm{j},p(\tens{W})}\!\left[\left(\frac{\partial h_{\bm{j}}^{(l)}}{\partial h^{(l')}_{\bm{i}} }\right)^2\right] = \frac{1}{Z} \sum_{\bm{j}} \sum_{\bm{C} \in \mathbb{P}^{l,l'}_{\bm{i},\bm{j}}} \prod_{l \in \mathbb{L}[\bm{C}]}
    \mathbb{V}\left[  w^{(l)}\right]
\end{equation}
where $\mathbb{P}^{l,l'}_{\bm{i},\bm{j}}\equiv\mathbb{P}_{h_{\bm{j}}^{(l)} \to h_{\bm{i}}^{(l')}}$ as defined in equation~\ref{eq:perm} and $p(\tens{W})$ is the distribution of the network weights and $\mathbb{L} \equiv Layers(\bm{C})$ is an operator that returns the sequence of layers associated with the commands (i.e. the layer reached by the previous command).
\end{proposition}

The proof of Proposition~\ref{propo:restricted} is given in Appendix~B.

Proposition~\ref{propo:restricted} has the interesting consequence that the expected \FeatureImportance is independent of the input unit value distribution and of the convolutional layer biases if present.
Also, if all paths cross the same set of layers (similarly to what can be seen in VGG architectures \cite{Simonyan2014}), then the squared entry importance is proportional to the product of the number of paths ending on this features times the product of the weight variances:
\begin{equation}
    \mathbb{E}_{\bm{j},p(\tens{W})}\!\!\left[\!\left(\frac{\partial h_{\bm{j}}^{(l)}}{\partial h^{(l')}_{\bm{i}} }\right)^2\right]\!=\!\frac{1}{Z}\!\prod_{l=0}^{L-1}\!
    \mathbb{V}\!\left[  w^{(l)}\right] \times \# \overline{p}_{\tens{H}^{(l)} \to h_{\bm{i}}^{(l')}}
\end{equation}

Showing that this result generalises to test-time batch normalisation \cite{Ioffe2015} and residual connections is straightforward. 
Regarding the former, the fact that the $BatchNorm$ operation is affine at test time ensures that Equation~\ref{eq:importance_paths} is still valid, up to a multiplicative constant reflecting the $BatchNorm$ weight multiplication and the ``offline'' z-scoring.
During training however, the variance of the partial derivatives of the network output are not independent of $h_{\bm{i}}^{(l')}$ anymore, due to the ``online'' z-scoring.
In Appendix~C, we show that under specific conditions, the expected importance landscape of a convolutional CNN with $BatchNorm$ layers depends upon the network topology (that is, the number of paths) and the variance of the input elements, and \textit{not} upon the weight variance anymore.


Regarding the latter, for many architectures, the effect of residual connections is simply to add a constant number of identical sub-paths from output to input in the network, regardless of which unit location is being considered. 
In other words, in such conditions, residual connections do not change the fact that the squared input entry salience is monotonically and linearly conditioned on the number of output-to-input paths that end up on it.
In situations where residual connections act as downsampling layers with a stride greater than one, they do indeed modify the number of paths for a subset of locations and not for another, making some parts of the input tensors more salient than others.

\subsection{Correcting for the number of paths}\label{sec:solution}
\comments{Questions for Sim and Reza: do you think that a figure, like the ones I have used in the all hands deck, are necessary? Since we will be, as usual, short on space I wouldn't use them unless they are truly necessary.}

We have just shown that, after initialisation, each pixel importance is on average determined by the network topology.
In this section, we detail how one can derive a regularisation loss for CNNs in order to make the importance landscape less sensitive to the network topology during training.

It should be emphasised beforehand that several existing data augmentation methods indirectly help the networks to be less sensitive to the feature location in the training dataset. 
Specifically, random image modifications (cropping, zooming, flipping, noising etc.) present the training images in various ways that have the effect of placing the relevant features at different locations. 
Hence, CNNs trained in such a way across multiple epochs are being presented with identical features with variable locations (or masks in the case of noise adjunction), making them hopefully oblivious to this factor at test time. 
We suggest that the fact that these modifications help the networks to generalize better can be seen as indirect evidence for the claim that, with non-trivial datasets, translation invariance of CNNs is not verified.
Nevertheless, CNNs will usually be biased towards some coordinates in the input images and this can harm the training process.

For simplicity, let us consider the simple case of a mini-batch consisting of a single element.
We also assume that the network is used in ``evaluation'' mode, that is, the $BatchNorm$ layers do not keep track of the running input sufficient statistics and use the pre-recorded running sufficient statistics for inference.
In classification settings, the network outputs a vector $\bm{y} \in \mathbb{R}^D$ (or more precisely a tensor with $B=1$, $\textbf{d}_y=\textbf{1}$ and $C_{\text{out}}=D$) of real values that represents the pre-softmax $D$-class probability:
\begin{equation}\label{eq:softmax}
\begin{aligned}
    p(\bm{y} \mid \tens{x}) &\equiv 
    \softmax{\left(\bm{h}^{(L)}\right)}
\end{aligned}
\end{equation}
where $\softmax(\bm{h}) \triangleq \left\langle \frac{\exp{h^{(L)}_i}}{\sum_j \exp{h^{(L)}_j}}\right\rangle_{i=1}^{D}$.

We define the square-feature importance landscape of such a network as the following operator
\begin{align*}
    I_{\bm{i}}(f)^2 &\triangleq \mathbb{E}_{p(y,\tens{X})}\left[\left( \frac{\partial h^{(L)}_y }{\partial x_{\bm{i}}} \right)^2\right]\\
    &= \left.\frac{1}{M} \sum_{m=1}^M \sum_{y \in \bm{y}} p(y \mid \tens{X}_m) \left( \frac{\partial h^{(L)}_y }{\partial x_{\bm{i},m}} \right)^2 \right | \tens{X}_m \in p(\tens{X})
\end{align*}
where $f \equiv \bm{h}^{(L)}(\tens{X})$, $\bm{i}$ is the input element location, $h_y$ is the $y^{\text{th}}$ entry of the output vector $\bm{h}$, and $M$ is the sample size. 
Note that this operator is implicitly conditioned on the training data distribution, which \textit{does} have an impact on the landscape at test time for \textit{trained} CNNs (since the \textit{i.i.d.} assumptions made in Section~\ref{sec:methods} do not hold anymore).

In practice, what we wish is to measure how a 2D location is sensitive to the network topology, hence we use the following transformation of $I_{\bm{i}}(f)^2$:
\begin{align}\label{eq:2dlandscape}
    I'_{i_{\text{w}}, i_{\text{h}}}(f) 
    &=\sqrt{d_{\text{w}} d_{\text{h}}}\frac{\norm{\textbf{I}_{i_{\text{w}}, i_{\text{h}}}(f)^2}}{\norm{\textbf{I}(f)^2}}
\end{align}
where $i_{\text{w}} \in 1:d_{\text{w}}$ and $i_{\text{h}} \in 1:d_{\text{h}}$ are respectively the width and height coordinates, where the norm in the numerator is taken over the channel dimension, and where the norm of the denominator is taken to be the Frobenius norm. It can easily be checked that the landscape operator $\textbf{I}'(f)$ is one on expectation for every image.

The operator \[\textbf{I}'(f) \equiv \left[ \begin{array}{ccc}
     I'_{1,1}(f) & \dots & I'_{1,d_{\text{h}}}(f) \\
     \vdots & \ddots & \vdots \\
     I'_{d_{\text{w}},1}(f) & \dots & I'_{d_{\text{w}},d_{\text{h}}}(f)
\end{array}\right]\] outputs a $2D$ map of the pixel-wise importance in a CNN $f$ that we wish to optimize, which leads us to derive a general family of regularised objectives:
\begin{equation}\label{eq:pilcro}
\begin{aligned}
    \boldsymbol{\theta}^* \triangleq \argmin_{\boldsymbol{\theta}} &\frac{1}{M} \sum_{m}^M \text{CE}\left(f_{\boldsymbol{\theta}}\left(\tens{X}_m\right), \overline{y}_m\right)+\alpha \mathcal{L}\left(\bm{I}'(f_{\boldsymbol{\theta}})\right)
\end{aligned}
\end{equation}
where CE stands for Cross-Entropy loss, and for a given dataset of input-class pairs $\{(\tens{X}_m, \overline{y}_m)\}_{m=1}^M$ with $\overline{y}_m$ being the true class label, network parameters $\boldsymbol{\theta} \equiv \{\tens{W}, \tens{b}\}$ and where $\alpha$ is a regularisation hyperparameter.
We name this loss \P-objective (or \P-regularisation), where \P\,stands for \textbf{PILCRO}: \textbf{P}ixel-wise \textbf{I}mportance \textbf{L}andscape \textbf{C}urvature \textbf{R}egularised \textbf{O}bjective.

We identify two forms that the landscape loss function $\mathcal{L}\left(\bm{I}'(f_{\boldsymbol{\theta}})\right)$ can take, but many other could be derived as well.
A first method could be to directly minimise the curvature of this landscape. 
A second method consists in minimising the distance between the gradient landscape and a measure of information content in the training dataset.

\subsubsection{Curvature minimisation (\P$_c$)}
For simplicity, the only curvature measure we consider here is the finite-difference $k$-Laplacian operator $\Delta_k \bm{I}'(f)$  \cite{Pertuz2013} whose energy (i.e. empirical variance) $\hat{\sigma}^2\left[\Delta_k \bm{I}'(f)\right]$ is a real positive value indicating how sharp that landscape is. 
Computing the value of this metric is quite straightforward as it only involves performing a discrete convolution with kernel $k$ over the importance landscape, where $k$ is a hyperparameter. In this setting, the \P-regularization term takes the form
\begin{equation}\label{eq:pilcro_curv}
    \mathcal{L}_c\left(\bm{I}'(f_{\boldsymbol{\theta}})\right) \triangleq  \hat{\sigma}^2\left[\Delta_k \bm{I}'(f_{\boldsymbol{\theta}})\right].
\end{equation}

\subsubsection{Target landscape casting (\P$_t$)}
Although Equation~\ref{eq:pilcro_curv} smooths the importance landscape, it might still be desirable to inject \textit{a priori} knowledge about the location specific data content of the dataset at hand.
To do this, we precompute a measure of ``smoothness'' of each pixel location:
\begin{equation*}
    \widehat{S}_{i,j}(\tens{X}) =  \frac{1}{M} \sum_{m=1}^M\sqrt{d_{\text{w}} d_{\text{h}}} \frac{\norm{\left(\Delta_k \tens{X}_{m}\right)_{i_{\text{w}},i_{\text{h}}}}}{\norm{\left(\Delta_k \tens{X}_{m}\right)}} 
\end{equation*}
which is analogous to the 2D-feature importance landscape of Equation~\ref{eq:2dlandscape}. This naturally leads us to propose the following \P-regularization:
\begin{equation}\label{eq:pilcro_target}
    \mathcal{L}_{t}\left(\bm{I}'(f_{\boldsymbol{\theta}})\right) \triangleq  \norm{\bm{I}'(f_{\boldsymbol{\theta}})-\widehat{\bm{S}}(\tens{X})}_2^2.
\end{equation}
In practice, this loss ``casts'' the feature importance landscape to a target importance map retrieved from the data.

\begin{figure*}[t]
  \centering
  \includegraphics[width=0.85\textwidth]{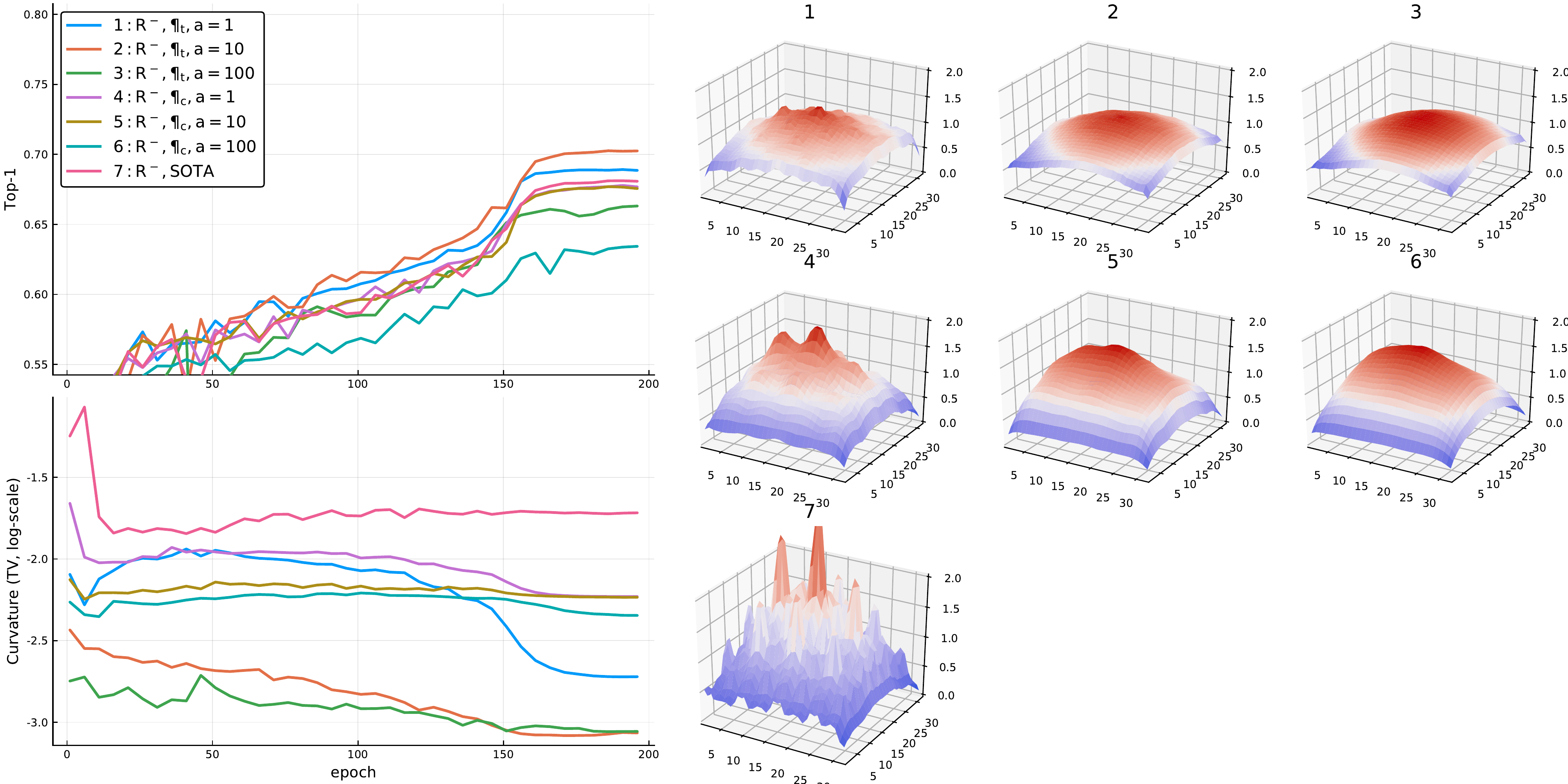}
  \caption{CIFAR-100 with ResNet-50 results in the RC$^-$ setting. The importance landscape heatmaps are plotted, numbers refer to the corresponding model (see the legend).}
  \label{fig:cifar_rc0}
\end{figure*}

\begin{figure*}[t]
  \centering
  \includegraphics[width=0.85\textwidth]{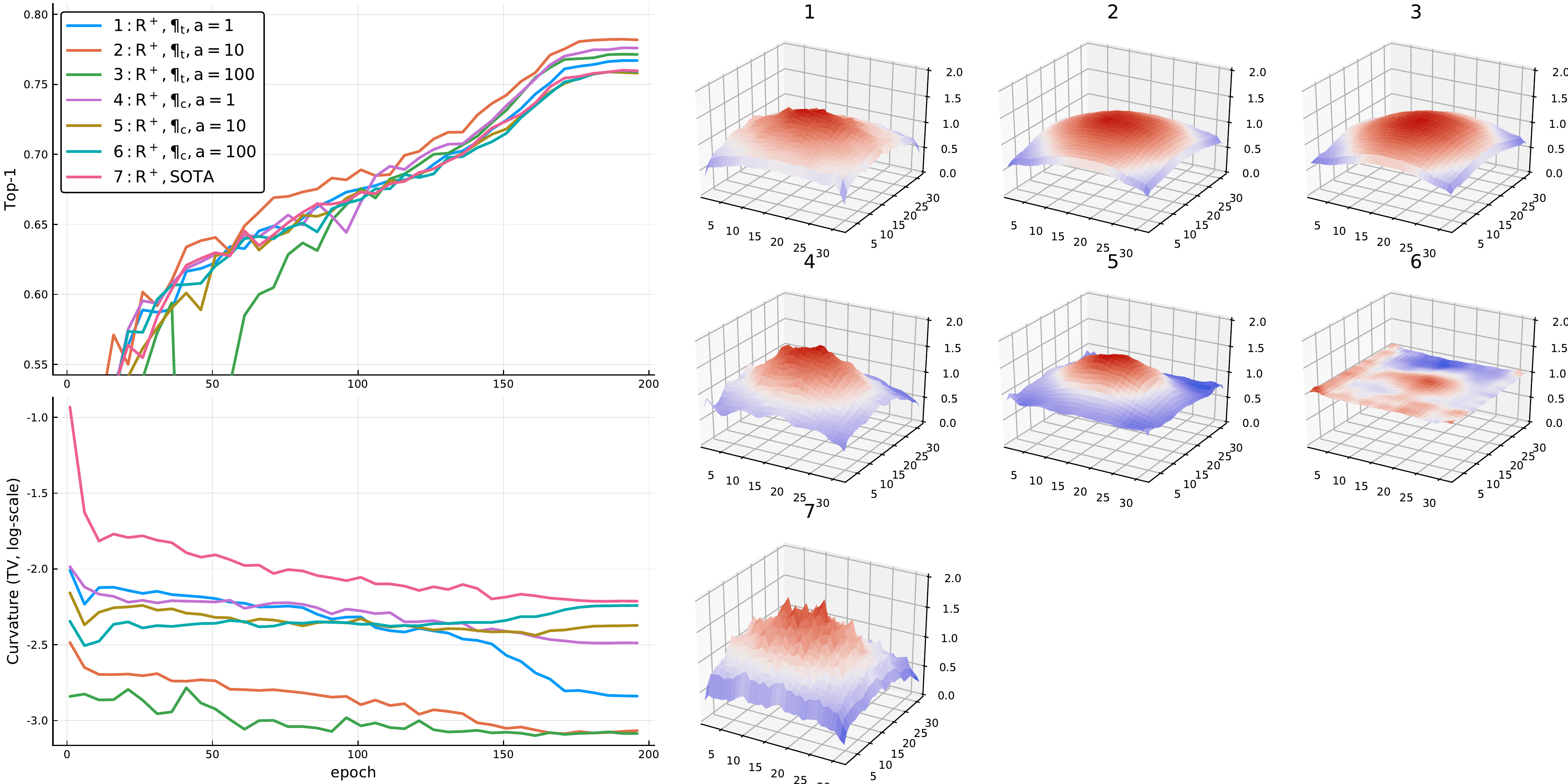}
  \caption{CIFAR-100 with ResNet-50 results in the RC$^+$ setting.}
  \label{fig:cifar_rc1}
\end{figure*}

\subsubsection{Practical considerations}
Two factors make the evaluation of Equation~\ref{eq:pilcro} laborious and need to be overcome: first, it relies on the estimation of the importance landscape $\bm{I}(f)^2$ which involves computing the square-jacobian vector product over the whole dataset.
Instead, we use the central limit theorem to get an approximation of this value using the probabilities defined in Equation~\ref{eq:softmax}:
\[
\bm{I}(f)^2 \approx \left.\frac{1}{B} \sum_{m=1}^B \left( \frac{\partial h^{(L)}_y }{\partial x_{\bm{i},m}} \right)^2 \right | \tens{X}_B \in p(\tens{X}), \, y \sim p(y \mid \tens{X}_m)
\]
which is true on expectation, and allows us to compute the regularisation loss online.

The second difficulty comes from the fact that computing the gradient of the \P-regularisation loss requires the importance landscape loss to be differentiated, which in turn necessitates two consecutive backward passes through the computational graph. 
To lower the associated computational burden, we use two distinct methods: first, we only compute the loss once every $T$ optimisation steps, and multiply the regularisation loss by a factor of $\alpha = \alpha_0 T$\footnote{which in practice amends to choosing a hyperparameter $\alpha$ that scales with $T$}. 
Second, when computing the regularisation loss, we switch the network to evaluation mode, which makes the second derivative of $f$ equal to zero almost everywhere, thereby reducing the memory and time complexity of this operation.

\section{Experiments}\label{sec:results}

\begin{table}
\small{
\centering
\begin{tabular}{ |l|l|llll| }
  \hline
  RC & Model & Top-1 & Top-5 & TV & Loss\\
  \toprule
  RC$^{-}$&SOTA                 &0.6825 &0.8918 &0.0192 & 1.4807\\
  RC$^{-}$&\P$_t$, $\alpha$=1   &0.6901 &0.8973 &0.00191 & 1.4241 \\
  RC$^{-}$&\P$_t$, $\alpha$=10  & \textbf{0.7041} & \textbf{0.9053} & \textbf{0.00094} & \textbf{1.3850} \\
  RC$^{-}$&\P$_t$, $\alpha$=100  &0.6653 &0.8844 &0.00097 & 1.5300 \\
  RC$^{-}$&\P$_c$, $\alpha$=1   &0.6788 &0.8938 &0.00591 & 1.4764 \\
  RC$^{-}$&\P$_c$, $\alpha$=10  &0.6792 &0.8859 &0.00585 & 1.5158\\
  RC$^{-}$&\P$_c$, $\alpha$=100 &0.6356 &0.8580 &0.00453 & 1.8577\\
  \hline
  RC$^{+}$&SOTA                 &0.7610 &0.9318 &0.00605 &1.3221\\
  RC$^{+}$&\P$_t$, $\alpha$=1   &0.7684 & 0.937 &0.00147 & 1.2664\\
  RC$^{+}$&\P$_t$, $\alpha$=10  & \textbf{0.7833} & \textbf{0.9431} &0.00088 & 1.2117\\
  RC$^{+}$&\P$_t$, $\alpha$=100  &0.7731 &0.9403 & \textbf{0.00086} & \textbf{1.1095}\\
  RC$^{+}$&\P$_c$, $\alpha$=1   &0.7773 &0.9405 &0.00327 &1.2107\\
  RC$^{+}$&\P$_c$, $\alpha$=10  &0.7597 &0.9306 &0.00422 &1.3011\\
  RC$^{+}$&\P$_c$, $\alpha$=100 &0.7604 &0.9312 &0.00567 &1.3709 \\
  \hline
\end{tabular}
\caption{CIFAR-100 results with a standard ResNet-50 architecture. Bold numbers indicate top-performing model for the considered metric.}
\label{tab:cifar}
}
\end{table}

We tested our regularized objective on two well known classification tasks: CIFAR-100 \cite{Krizhevsky09learningmultiple}, constituted of 50K training and 10K validation 32 $\times$ 32 images, and the 2012 ImageNet dataset \cite{ILSVRC15} with over a million medium to large size images for the training dataset and 50K images for the validation dataset. 
As a measure of performance, we used the Top-1 and Top-5 accuracy.
\paragraph{Hyperparameter settings}
In all experiments, we tested various values of $\alpha$ ($\alpha \in \{1, 10, 100\}$ for the \P$_c$-objective and \P$_t$-objective), whereas the \P-regularisation interval $T$ was set to 10 iterations in all experiments.
All discrete-Laplacian kernel width values were set to $k=3$.
Finally, we trained CIFAR-100 with (RC$^+$) and without (RC$^-$) random cropping to test for the model sensitivity to data augmentation.
The rest of the hyperparameters used are detailed in Appendix~D.

\subsection{Results}
\subsubsection{CIFAR-100 dataset}
In both RC settings, it was apparent that models with a lower curvature were performing better (Table~\ref{tab:cifar}), with a curvature measured as the total variation (TV, i.e. average square difference between nearby locations) of the empirical gradient landscape computed over the validation dataset. 
In particular, the \P$_t$-objective with $\alpha=10$ performed best on both random cropping settings, with an average performance increase of roughly 2\% (1\% for $\alpha=1$).

As expected, regularisation weight had a significant impact on performance: a completely flat landscape such as the one that can be observed for the RC$^{+}$, \P$_c$-objective with a high regularisation weight $\alpha=100$ (see Figure~\ref{fig:cifar_rc0} and Figure~\ref{fig:cifar_rc1}) did not perform better than the State-of-the-art (SOTA) configuration, whereas the same experimental setting with $\alpha=1$ outperformed SOTA.
This suggests that regularization was better achieved when it was guided by data.

\begin{table}
\small{\centering
\begin{tabular}{ |l|llll| }
  \hline
  Model & Top-1 & Top-5 & TV at Top-1 & Loss\\
  \toprule
  SOTA                 & 0.7768 & 0.9384 & 0.00499 & 1.4045 \\
  \P$_t$, $\alpha$=1   & 0.7780 & 0.9387 & 0.00048 & 1.4136\\
  \P$_t$, $\alpha$=10  & \textbf{0.7800} & 0.9385 & 0.00019 & 1.4022 \\
  \P$_t$, $\alpha$=100  & 0.7702 & 0.9350 & \textbf{0.00007} & \textbf{1.4017} \\
  \P$_c$, $\alpha$=1   & 0.7790 & 0.9374 & 0.00102 & 1.4072 \\
  \P$_c$, $\alpha$=10  & 0.7797 & \textbf{0.9398} & 0.00030 & 1.4046 \\
  \P$_c$, $\alpha$=100 & 0.7755 & 0.9366 & 0.00024 & 1.4097 \\
  \hline
\end{tabular}
\caption{ImageNet results on square 288 $\times$ 288 images.}
\label{tab:imagenet}
}
\end{table}

It also appeared that the use of random cropping was helping SOTA and regularized versions of the model to flatten and smooth the gradient norm landscape, hence corroborating the assertion previously made that data augmentation soften the bias caused by the architecture topology on the capability to attend evenly different parts of the input images.

To test the robustness of the various networks we trained, we applied a wide set of perturbation types divided in four main categories (Noise, Weather, Digital, Blur) as suggested by \cite{Hendrycks2019} across five different levels to all of them.
Results are detailed in Appendix~E. 
Overall, they strongly suggested that \P$_t$-objective with $\alpha=10$ models were the most robust configuration for the Weather, Digital and Blur categories, whereas SOTA instances outperformed other models for the Noise category.

\subsubsection{ImageNet dataset}
On ImageNet, \P-regularisation provided a modest gain in accuracy for both \P$_c$ and \P$_t$ methods, that was consistent across Top-1, Top-5 and Loss metrics (Table~\ref{tab:imagenet} and Figure~\ref{fig:imagenet_rc1} in Appendix~F).

To get a sense of the model robustness to data perturbation, we tested how vulnerable models were to ablation of relevant information.
In short, for each model, we computed the salience map using the technique of Integrated Gradient \cite{Sundararajan2017} for 512 randomly selected correctly classified images from the validation dataset, using 1000 other images of the dataset as baseline. Next, we sorted each of the pixels according to its relevance per image. Finally, we iteratively replaced the top-$K$ ($\forall K\in\{1,\dots,d_x^2\}$) pixels by their image and channel-wise average value.
For each of these new, noisy images, we measured the ratio of the target class probability change with respect to the baseline as a function of $K$.
This procedure shows of how sensitive a classifier is to its top-$K$ most important pixels: the more vulnerable to noise, the sharper the drop of accuracy.
As shown in Figure~\ref{fig:imagenet_robustness} in Appendix~F, it appeared again that the \P$_t$-objective with $\alpha=10$ trained model performed better than other networks for this metric, which was consistent with other metrics, such as curvature, loss and accuracy.

Like we did for CIFAR-100, we also tested robustness of the trained models to different image perturbations.
Although results were less clear than with CIFAR-100, we still observed a greater robustness for \P-regularised models than their naive counterparts, that was consistent per perturbation category and per \P-objective setting (Appendix~E).

\section{Conclusion}

This paper uncovers an apparently ignored issue in the ML community that has to do with how fairly a neural network treats various input locations, and how gradient flows in neural networks. We propose a regularisation loss (\P-objective) to deal with this issue.
Further work will be required to develop architectures that mitigate this effect as well as metrics that can measure its impact on the network performance and fairness.





\bibliography{pilcro}
\bibliographystyle{icml2020}

\onecolumn{
\appendix

\section{Proof of Lemma 1}

To show that in CNNs, the (squared) permutation-based measure of feature importance can be approximated by the squared expected output-to-input partial derivative, we first start by deriving the formula of the former in a model $f$ receiving two input units $\bm{x} \triangleq \langle x_0,x_1\rangle$:
\begin{align}\label{eq:PFI}
    I(x_0) &= \mathbb{E}_{p(\bm{x})}\left[\left( \mathbb{E}_{p(x_0)}\left[f(x_0, x_1)\right] - f(x_0, x_1) \right)^2 \right].
\end{align}
Next, we expand the inner expectation using its second order Taylor series:
\begin{equation}\label{eq:Taylor}
    \begin{aligned}
    \begin{split}
    &\mathbb{E}_{p(x_0)}\left[f(x_0, x_1)\right]\\
    &\quad=\mathbb{E}_{p(x_0)}\Bigl[f(\overline{x}_0, x_1) +\frac{\partial f(\overline{x}_0, x_1)}{\partial \overline{x}_0} (x_0 - \overline{x}_0) \\
    &\qquad+ \frac{1}{2}\frac{\partial^2 f(\overline{x}_0, x_1)}{\partial \overline{x}_0^2}(x_0 - \overline{x}_0)^2 + \dots \Bigr]\\
    &\quad= f(\overline{x}_0, x_1)+\frac{1}{2}\frac{\partial^2 f(\overline{x}_0, x_1)}{\partial x_0^2} \sigma(x_0)^2 + \dots.
    \end{split}
    \end{aligned}
\end{equation}
A curiosity of CNNs with $ReLU$ activation functions is that the second derivative of the (pre-loss) output layer \textit{w.r.t.} the input units is zero almost everywhere (\cite{Yao2018}), which simplifies the above equation to 
\begin{align}\label{eq:simpleTaylor}
    \mathbb{E}_{p(x_0)}\left[f(x_0, x_1)\right] &\approx  f(\overline{x}_0, x_1).
\end{align}
Plugging Equation~\ref{eq:simpleTaylor} into Equation~\ref{eq:PFI} now gives:
\begin{align}\label{eq:PFI2}
    I(x_0) &\approx \mathbb{E}_{p(\bm{x})}\left[\left( f(\overline{x}_0, x_1) - f(x_0, x_1)\right)^2 \right].
\end{align}
Considering the negative difference $\delta$ between $x_0$ and $\overline{x}_0$, we can substitute $\overline{x}_0 = x_0 + \delta$ which leads to 
\begin{align*}
    I(x_0) &\approx \mathbb{E}_{p(\bm{x})}\left[\left( f(x_0 + \delta, x_1) - f(x_0, x_1) \right)^2 \right].
\end{align*}
As the partial derivative of $f(x_0, x_1)$ \textit{w.r.t.} $x_0$ is defined as
\begin{align*}
    \frac{\partial f(x_0, x_1)}{\partial x_0} = \lim_{\delta \rightarrow 0}\frac{f(x_0 + \delta, x_1) - f(x_0, x_1)}{\delta},
\end{align*}
and under the assumption that $x_0$ and $\overline{x}_0$ are sufficiently close (common to Equation~\ref{eq:Taylor}), the following approximation holds:
\begin{align*}
    I(x_0) &\approx \mathbb{E}_{p(\bm{x})}\left[\delta^2 \left( \frac{\partial f(x_0, x_1)}{\partial x_0} \right)^2 \right]\\
    &\approx \mathbb{V}[x_0]\mathbb{E}_{p(\bm{x})}\left[\left( \frac{\partial f(x_0, x_1)}{\partial x_0} \right)^2 \right]
\end{align*}
where we have taken advantage of the fact that $\mathbb{E}\left[\delta^2\right] = \mathbb{V}[x_0]$ by definition of the variance. 
\qedwhite

\section{Proof of Proposition 1}
Proposition~1 follows directly from the chain rule:
\begin{align}\label{eq:importance_path_naive}
    \mathbb{E}\!\left[\left(\frac{\partial h_{\bm{j}}^{(l)}}{\partial h^{(l')}_{\bm{i}} }\right)^2\right]\!&=
    \mathbb{E}\!\left[\left( \sum_{\bm{C} \in \mathbb{P}^{l,l'}_{\bm{i},\bm{j}}} \prod_{l \in \mathbb{L}[\bm{C}]} w^{(l)}_{\bm{c}^{(l)}}\right)^2\right]
\end{align}
and from the fact that, under most initialisation methods, including Glorot \cite{Glorot2010} and He\linebreak \cite{He2015}, the weights of the network are zero on expectation. This simplifies the above equation significantly, as the expectation of the product of two independent draws of a same distribution with zero mean is zero, that is:
\begin{align*}
    &\mathbb{E}_{\bm{W}_i \sim p(\bm{W}), \bm{W}_j \sim p(\bm{W})} \left[ \bm{W}_i \bm{W}_j  \right] = \0\\
    &\quad \text{if}\,  \mathbb{E}_{p(\bm{W})}[\bm{W}_i] = \0\,\text{for all}\,i.
\end{align*}
Hence,
\begin{align}\label{eq:importance_paths}
    \mathbb{E}_{\bm{j}, p(\tens{W})}\!\left[\left(\frac{\partial h_{\bm{j}}^{(l)}}{\partial h_{\bm{i}}^{(l')} }\right)^2\right]\!&=\!\!\sum_{\bm{C} \in \mathbb{P}^{l,l'}_{\bm{i},\bm{j}}} \prod_{l \in \mathbb{L}[\bm{C}]}
    \mathbb{E}_{p\left(w^{(l)}\right)}\left[\left( w^{(l)}_{\bm{c}^{(l)}}\right)^2\right]
\end{align}
which leads to Equation~\ref{eq:prop} by definition of the variance. \qedwhite

\section{Effect of nonlinearities}\label{sec:app:NonLinearities}
Nonlinear activation functions and other operations (typically normalization at train time) break the above logic and make the expected squared output-input partial derivative dependent on other factors than the number of paths and the weights variance.
This means that the expected feature importance is not independent of the data anymore, and the picture is therefore more complex.

Here, we focus on the Restricted Linear Unit (ReLU) activation function (\cite{Nair2010}).
Consider the squared partial derivative of the operation $h^{(l)} = ReLU\left(a^{(l)}\right)$:
\begin{align}
\left(\frac{\partial h^{(l)}}{\partial a^{(l)}}\right)^2 = \left \{ \begin{array}{ll} 
1 & \text{if } a^{(l)} > 0\\
0 & \text{otherwise}
\end{array}
\right.
\end{align}
which is on expectation equal to $\frac{1}{2}$ if $a^{(l)}$ is equal to $0$ on expectation.
For later, it will be interesting to remember that, assuming that $\tens{A}^{(l)}$ entries are \textit{i.i.d.} and follow a normal distribution $\mathbb{N}(0, S)$, the output variance is equal to
\begin{equation*}
    \mathbb{V}\left[h^{(l)}\right] = \frac{\mathbb{V}_{x \sim \mathbb{N}^+}[x]}{2} - \left(\frac{\mathbb{E}_{x \sim \mathbb{N}^+}[x]}{2}\right)^2
\end{equation*}
where $\mathbb{N}^+$ is a $[0, \infty)$ truncated normal distribution.

To some degree, the case of $BatchNorm$ layers is harder to handle.
Training time batch normalization and its derivatives essentially entail a z-scoring of the input value -- most often the output of a convolutional layer -- over some dimension of the input tensor, typically the batch, height and width dimension for two-dimensional CNNs.
The following lemma shows how $BatchNorm$ layers typically impact the \FeatureImportance in a CNN:
\begin{lemma}\label{lemma:bn_importance}
In a neural network where convolutions are interleaved with batch normalization layers, the salience of a given input entry is in good approximation equal to
\begin{equation}\label{eq:definite_bn_importance}
\begin{aligned}
    &\mathbb{E}_{\bm{j},\tens{X},\tens{W},\bm{\gamma}}\left[\left( \frac{\partial y_{\bm{j}} }{\partial x_{\bm{i}}} \right)^2\right] = \prod_{l=1}^{(L)}\frac{N^{(l)}-1}{N^{(l)}}\mathbb{E}_{{\gamma^{(l)}}}\left[{\gamma^{(l)}}^2\right] \mathbb{E}_{\bm{j}}\left[\frac{\# p_{a^{(l)}_{\bm{j}}\rightarrow h^{(l-1)}_{\bm{i}}}}{\argmax_{\bm{i}}\mathbb{E}_{\bm{j}'}\left[\# p_{a^{(l)}_{\bm{j}'}\rightarrow h^{(l-1)}_{\bm{i}}}\right]}\right]\\&\hspace{11em}\times\frac{1}{ \mathbb{V}\left[h^{(l-1)}\right]}+ \mathcal{O}\left(\sum_{l=1}^{(L)} {N^{(l)}}^{-1}\right)
\end{aligned}
\end{equation}
which is independent of the weight variance.
\end{lemma}
\textbf{Proof}:
The expected squared partial derivative of the output of a convolutional layer is equal to:
\begin{align*}
    &\mathbb{E}_{\bm{j}, \tens{W}, \tens{H}}\left[\left(\frac{\partial a^{(l+1)}_{\bm{j}}}{\partial h^{(l)}_{\bm{i}}}\right)^2\right] = \frac{1}{Z} \mathbb{V}\left[w^{(l)}\right] \mathbb{E}_{\bm{j}}\left[\# p_{a^{(l+1)}_{\bm{j}} \rightarrow h^{(l)}_{\bm{i}}}\right], 
    \\&\qquad \text{with } C_{\text{in}} \leq \mathbb{E}_{\bm{j}}\left[\# p_{a^{(l+1)}_{\bm{j}} \rightarrow h^{(l)}_{\bm{i}}}\right] \leq C_{\text{in}}  \prod_{k} \kappa_i
\end{align*}

The partial derivative of $BatchNorm$ layers takes two generic forms, 
\[
\frac{\partial BatchNorm(a^{(l)}_{\bm{j}}, \tens{A}^{(l)})}{\partial a^{(l)}_{\bm{i}}} = \left\{\begin{array}{ll}
     \gamma \frac{N^{(l)}-1}{N^{(l)} \sigma(\bm{A}_{c_{\text{out}}}^{(l)})} - \frac{\left(a_{\bm{i}} - \overline{\bm{A}_{c_{\text{out}}}^{(l)}}\right)^2}{N \sigma\left(\bm{A}_{c_{\text{out}}}^{(l)}\right)^3}& \text{ if } j = i \\
     \gamma \frac{-1}{N^{(l)} \sigma\left(\bm{A}_{c_{\text{out}}}^{(l)}\right)}  - \frac{\left(a_{\bm{i}} - \overline{\bm{A}_{c_{\text{out}}}^{(l)}}\right)^2}{N \sigma\left(\bm{A}_{c_{\text{out}}}^{(l)}\right)^3}& \text{ otherwise}
\end{array}\right.
\]
from which we can derive the squared derivative:
\begin{equation}\label{eq:sq_deriv_bn}
    \begin{aligned}
\left(\frac{\partial BatchNorm(a^{(l)}_{\bm{j}}, \tens{A}^{(l)})}{\partial a^{(l)}_{\bm{i}}}\right)^2 &= 
\left\{\begin{array}{ll}
     \gamma^2 \left(\frac{N^{(l)}-1}{N^{(l)}}\right)^2 \frac{1}{\sigma\left(\bm{A}^{(l)}_{c_{\text{out}}}\right)^2} + \left(\frac{\left(a_{\bm{i}} - \overline{\bm{A}_{c_{\text{out}}}^{(l)}}\right)^2}{N \sigma(\bm{A}^{(l)}_{c_{\text{out}}})^3}\right)^2\span\\
     \qquad - 2 \frac{N-1}{N^{(l)} \sigma(\bm{A}^{(l)}_{c_{\text{out}}})} \frac{\left(a_{\bm{i}} - \overline{\bm{A}_{c_{\text{out}}}^{(l)}}\right)^2}{N \sigma(\bm{A}^{(l)}_{c_{\text{out}}})^3} & \text{ if } j = i \\
     \gamma^2 \left(\frac{-1}{N^{(l)}}\right)^2 \frac{1}{\sigma\left(\bm{A}^{(l)}_{c_{\text{out}}}\right)^2} + \left(\frac{\left(a_{\bm{i}} - \overline{\bm{A}_{c_{\text{out}}}^{(l)}}\right)^2}{N \sigma(\bm{A}^{(l)}_{c_{\text{out}}})^3}\right)^2\span\\
     \qquad - 2 \frac{-1}{N^{(l)} \sigma(\bm{A}^{(l)}_{c_{\text{out}}})} \frac{\left(a_{\bm{i}} - \overline{\bm{A}_{c_{\text{out}}}^{(l)}}\right)^2}{N \sigma(\bm{A}^{(l)}_{c_{\text{out}}})^3} & \text{ otherwise}
\end{array}\right.\\
&=\left\{\begin{array}{ll}
     \gamma^2 \left(\frac{N^{(l)}-1}{N^{(l)}}\right)^2 \frac{1}{\sigma\left(\bm{A}^{(l)}_{c_{\text{out}}}\right)^2} + \mathcal{O}\left(N^{(l)}\right)^{-2} & \text{ if } j = i \\
     \mathcal{O}\left(\left(N^{(l)}\right)^{-2}\right) & \text{ otherwise.}
\end{array}\right.
\end{aligned}
\end{equation}
Equation~\ref{eq:sq_deriv_bn} shows that we can derive upper bound the squared derivative of the $BatchNorm$ layer, which is tight when the sample size (typically equal to the product of the batch size, the width and the height of the layer considered) is high.

Let us now derive the expected value of the squared partial derivative of the $BatchNorm \circ Conv$ composite function:
\begin{align*}
    &\mathbb{E}_{\bm{j},\tens{H}^{(l-1)},\tens{W}^{(l-1)},\gamma}\left[\left( \frac{\partial \left. BatchNorm\left(a_{\bm{j}}^{(l)}, \tens{A}^{(l)}\right) \right|_{\tens{A} = \tens{W}^{(l-1)} * \tens{H}^{(l-1)}} }{\partial h^{(l-1)}_{\bm{i}}} \right)^2\right] \approx\\
    &\quad\mathbb{E}_{\bm{j},\tens{H}^{(l-1)},\tens{W}^{(l-1)},\gamma}\left[ \gamma^2 \left(\frac{N^{(l)}-1}{N^{(l)}}\right)^2 \frac{1}{\sigma\left(\bm{A}^{(l)}_{c_{\text{out}}}\right)^2} \left(\frac{\partial a_{\bm{j}}}{h_{\bm{i}^{(l-1)}}}\right)^2  \right]
\end{align*}
where the expectation is taken \textit{w.r.t.} the weight distribution and the input data.
We lower bound this expression using Jensen's inequality for the inverse variance expectation
\begin{equation*}
    \begin{aligned}
        \mathbb{E}\left[\frac{1}{\sigma(x)^2}\right] &\geq \frac{1}{\mathbb{E}\left[\sigma(x)^2\right]}\\
        &=\frac{N}{N-1} \frac{1}{\mathbb{V}[x]},
    \end{aligned}
\end{equation*} 
which has an error of the order $\mathcal{O}\left({\left(N^{(l)}\right)^{-1}}\right)$, again making this approximation tight for large $N^{(l)}$:
\begin{equation}\label{eq:exp_sq_composite}
\begin{aligned}
    \mathbb{E}_{\bm{j},\tens{H}^{(l-1)},\tens{W}^{(l-1)},\gamma}\left[\left( \frac{\partial \left. BatchNorm\left(a_{\bm{j}}^{(l)}, \tens{A}^{(l)}\right) \right|_{\tens{A} = \tens{W}^{(l-1)} * \tens{H}^{(l-1)}} }{\partial h^{(l-1)}_{\bm{i}}} \right)^2\right] \approx
    \\\quad \mathbb{E}_{\bm{j},\tens{H}^{(l-1)},\tens{W}^{(l-1)},\gamma}\left[ \gamma^2 \frac{N^{(l)}-1}{N^{(l)}} \frac{1}{\mathbb{V}\left[\bm{A}^{(l)}_{c_{\text{out}}}\right]} \left(\frac{\partial a_{\bm{j}}}{h_{\bm{i}^{(l-1)}}}\right)^2\right].
\end{aligned}
\end{equation}

Using the rule of the total expectation, and noting that under the assumption that the entries of $\tens{H}^{(l-1)}$ are \textit{iid.}, the variance of the output of the convolution is equal to 
\[
\mathbb{V}_{\tens{H}^{(l-1)}}\left[\bm{A}^{(l)}_{c_{\text{out}}}\right] = \mathbb{V}\left[h^{(l-1)}\right] \sum_k \sum_{c_{\text{in}}}
\left(w^{(l-1)}_{k,c_{\text{out}},c_{\text{in}}}\right)^{2}
\]
Equation~\ref{eq:exp_sq_composite} can be decomposed as
\begin{equation}\label{eq:bn_effect}
\begin{aligned}
    &\mathbb{E}_{\bm{j},\tens{H}^{(l-1)},\tens{W}^{(l-1)},\gamma}\left[\left( \frac{\partial \left. BatchNorm\left(a_{\bm{j}}^{(l)}, \tens{A}^{(l)}\right) \right|_{\tens{A} = \tens{W}^{(l-1)} * \tens{H}^{(l-1)}} }{\partial h^{(l-1)}_{\bm{i}}} \right)^2\right]
\\&\quad \approx \mathbb{E}_{\bm{j},\tens{W}^{(l-1)},\gamma}\left[  \gamma^2 \frac{N^{(l)}-1}{N^{(l)}} \frac{1}{ \mathbb{V}\left[h^{(l-1)}\right] \sum_k \sum_{c_{\text{in}}}
\left(w^{(l-1)}_{k,c_{\text{out}},c_{\text{in}}}\right)^{2} } \left(\frac{\partial a_{\bm{j}}}{h_{\bm{i}^{(l-1)}}}\right)^2  \right]\\
&\quad = \mathbb{E}_{\bm{j},\tens{W}^{(l-1)},\gamma}\left[  \gamma^2 \frac{N^{(l)}-1}{N^{(l)}} \frac{1}{ \mathbb{V}\left[h^{(l-1)}\right] \sum_k \sum_{c_{\text{in}}}
\left(w^{(l-1)}_{k,c_{\text{out}},c_{\text{in}}}\right)^{2} } \left(w^{(l-1)}_{\bm{k}}\right)^2  \right]\\
&\quad=\mathbb{E}_{\bm{j},\tens{H}^{(l-1)},\tens{W}^{(l-1)},\gamma}\left[ \gamma^2 \frac{N^{(l)}-1}{N^{(l)}} \frac{1}{ \mathbb{V}\left[h^{(l-1)}\right] \argmax_{\bm{i}}\mathbb{E}_{\bm{j}'}\left[\# p_{a^{(l)}_{\bm{j}'}\rightarrow h^{(l-1)}_{\bm{i}}}\right] } \# p_{a^{(l)}_{\bm{j}}\rightarrow h^{(l-1)}_{\bm{i}}}  \right]\\
&\quad=\frac{N^{(l)}-1}{N^{(l)}}\mathbb{E}_{\gamma}\left[\gamma^2\right] \mathbb{E}_{\bm{j}}\left[\frac{\# p_{a^{(l)}_{\bm{j}}\rightarrow h^{(l-1)}_{\bm{i}}}}{\argmax_{\bm{i}}\mathbb{E}_{\bm{j}'}\left[\# p_{a^{(l)}_{\bm{j}'}\rightarrow h^{(l-1)}_{\bm{i}}}\right]}\right]\frac{1}{ \mathbb{V}\left[h^{(l-1)}\right]} 
\end{aligned}
\end{equation}
where $\bm{k}$ is the index of the weight entry \textit{s.t.} $a_{\bm{j}} = w^{(l-1)}_{\bm{k}} h^{(l-1)}_{\bm{i}}$.
It becomes apparent in Equation~\ref{eq:bn_effect} that $BatchNorm$ layers cancel the effect of the variance of convolutional layers, ensuring that the expected squared partial derivative of the output layer is in good approximation dependent of the input variance only.

The generalization to multiple layers follows by simple recursivity.

\qedwhite

As we can see, $BatchNorm$ layers decouple the variance of the weights of the discrete convolutions from the measure of input unit importance, at the cost of making it a function of the input variance.

\section{Hyperparameter settings}

\subsection{CIFAR-100 dataset}
We tested the CIFAR-100 dataset on the ResNet50 architecture (\cite{He2016}) over a single random seed.
The same standard training hyperparameters were used for all optimisations: batch size of 128 samples, SGD optimizer with momentum set with an initial learning rate of 0.1 decaying with a cosine schedule over 200 epochs. 
We experimented two data augmentation settings where random horizontal flipping was (RC$^{+}$) or was not (RC$^{-}$)  accompanied by 0 to 4 pixels random-cropping with reflection-mode padding.
A weight decay of $10^{-5}$ was used according to SOTA standards. 

\subsection{ImageNet dataset}
We tested ResNet50 on the 2012 ImageNet dataset with a standard hyperparameter configuration: batch size of 256 images, SGD with initial learning rate of 0.256 and cosine schedule, momentum of 0.875, label smoothing of 0.1, weight decay of $3 \times 10^{-5}$ that excluded $BatchNorm$  trainable parameters. Data augmentation included random zooming, and images were cropped with 224 pixels square input dimension.
Because of the absence of random cropping for ImageNet, we did not include RC$^{+/-}$ experiments in this section. 
Results are reported over a single random seed.

\begin{landscape}
\section{Perturbation robustness}\label{sec:app:robustness}
\begin{table}[h]
\small{
\begin{tabular}{c|ccccc|ccccc|ccccc|ccccc}
Model&\multicolumn{5}{c}{Noise}&\multicolumn{5}{c}{Blur}&\multicolumn{5}{c}{Weather}&\multicolumn{5}{c}{Digital}\\
\toprule
&1&2&3&4&5&1&2&3&4&5&1&2&3&4&5&1&2&3&4&5\\
\midrule
RC$^-$, \P$_t$, $\alpha=$1 &0.36&0.18&0.08&0.05&0.04&0.66&0.58&0.51&0.42&0.34&0.85&0.70&0.64&0.59&0.45&0.79&0.67&0.59&0.48&0.34\\
RC$^-$, \P$_t$, $\alpha=$10 &0.51&0.29&0.15&0.10&0.07&\textbf{0.67}&\textbf{0.59}&0.52&0.43&0.34&\textbf{0.86}&\textbf{0.72}&\textbf{0.66}&\textbf{0.61}&\textbf{0.47}&\textbf{0.81}&\textbf{0.69}&\textbf{0.61}&\textbf{0.50}&\textbf{0.37}\\
RC$^-$, \P$_t$, $\alpha=$100 &0.40&0.20&0.09&0.06&0.04&0.66&0.59&\textbf{0.52}&\textbf{0.43}&\textbf{0.35}&0.86&0.70&0.63&0.57&0.43&0.79&0.67&0.59&0.47&0.36\\
RC$^-$, \P$_c$, $\alpha=$1 &0.40&0.22&0.12&0.08&0.06&0.65&0.56&0.47&0.39&0.30&0.84&0.69&0.62&0.56&0.41&0.77&0.65&0.57&0.46&0.33\\
RC$^-$, \P$_c$, $\alpha=$10 &0.24&0.11&0.05&0.04&0.03&0.65&0.57&0.49&0.40&0.32&0.84&0.70&0.63&0.57&0.43&0.78&0.65&0.58&0.47&0.34\\
RC$^-$, \P$_c$, $\alpha=$100 &0.39&0.24&0.13&0.09&0.06&0.64&0.55&0.47&0.38&0.32&0.81&0.66&0.59&0.53&0.39&0.75&0.60&0.54&0.43&0.32\\
RC$^-$, SOTA&\textbf{0.59}&\textbf{0.41}&\textbf{0.25}&\textbf{0.17}&\textbf{0.12}&0.66&0.58&0.49&0.41&0.32&0.85&0.70&0.63&0.58&0.43&0.78&0.67&0.60&0.49&0.36\\
\midrule
RC$^+$, \P$_t$, $\alpha=$1 &0.54&0.33&0.18&0.12&0.09&0.65&0.57&0.49&0.42&0.33&0.86&0.71&0.65&0.60&0.45&0.82&0.71&0.63&0.50&0.36\\
RC$^+$, \P$_t$, $\alpha=$10 &0.52&0.30&0.15&0.09&0.07&\textbf{0.67}&\textbf{0.60}&\textbf{0.52}&\textbf{0.44}&0.35&\textbf{0.87}&\textbf{0.74}&\textbf{0.68}&\textbf{0.63}&\textbf{0.49}&\textbf{0.84}&\textbf{0.74}&\textbf{0.66}&\textbf{0.55}&\textbf{0.39}\\
RC$^+$, \P$_t$, $\alpha=$100 &0.38&0.18&0.07&0.04&0.03&0.67&0.59&0.51&0.42&0.34&0.87&0.72&0.67&0.61&0.47&0.83&0.73&0.65&0.52&0.38\\
RC$^+$, \P$_c$, $\alpha=$1 &0.53&0.32&0.17&0.11&0.07&0.66&0.58&0.50&0.42&0.33&0.86&0.72&0.66&0.61&0.45&0.83&0.72&0.64&0.52&0.37\\
RC$^+$, \P$_c$, $\alpha=$10 &0.35&0.18&0.09&0.06&0.04&0.65&0.57&0.49&0.42&0.33&0.86&0.71&0.66&0.60&0.45&0.82&0.72&0.63&0.51&0.37\\
RC$^+$, \P$_c$, $\alpha=$100 &0.19&0.08&0.03&0.02&0.02&0.67&0.59&0.52&0.43&\textbf{0.35}&0.86&0.73&0.67&0.61&0.47&0.83&0.72&0.64&0.51&0.38\\
RC$^+$, SOTA&\textbf{0.59}&\textbf{0.39}&\textbf{0.23}&\textbf{0.16}&\textbf{0.11}&0.66&0.58&0.50&0.41&0.33&0.86&0.72&0.66&0.60&0.44&0.83&0.72&0.64&0.52&0.38\\
\end{tabular}
\caption{Image perturbation robustness of CIFAR-100 models. 
Measures are retrieved over the entirety of the validation dataset.
Per-category values correspond to average robustness measure over perturbation types as proposed by \cite{Hendrycks2019}. 
Robustness is defined as the average of the fraction $\frac{p_{c_{\text{max}}}^{\text{perturbed}}}{p^*_{c_{\text{max}}}}$, where $p$ is the class probability with ($^\text{perturbed}$) and without ($^*$) perturbation, $c_{\text{max}}$ is the class predicted by the classifier when no perturbation is applied.
Columns are divided in perturbation level, with 1 being the lowest level of perturbation, and 5 the highest.
Models with lowest sensitiveness to perturbations are indicated with bold characters.
The observation that the regularized models were more vulnerable to the ``noise'' perturbation indicates that uncorrelated pixel alterations had a qualitatively different impact on the prediction than correlated ones, such as Gaussian noise. 
}

\vspace{1em}
\begin{tabular}{c|ccccc|ccccc|ccccc|ccccc}
Model&\multicolumn{5}{c}{Noise}&\multicolumn{5}{c}{Blur}&\multicolumn{5}{c}{Weather}&\multicolumn{5}{c}{Digital}\\
\toprule
&1&2&3&4&5&1&2&3&4&5&1&2&3&4&5&1&2&3&4&5\\
\P$_t$, a=1 &0.61&\textbf{0.44}&\textbf{0.26}&\textbf{0.10}&0.03&0.58&0.41&0.23&0.15&0.10&0.65&0.47&0.40&0.32&0.23&0.72&0.57&0.48&0.29&0.14\\
\P$_t$, a=10 &0.59&0.40&0.22&0.09&0.03&0.58&0.42&0.24&0.15&0.10&0.65&0.48&0.40&0.33&0.23&0.73&0.59&0.51&0.32&\textbf{0.19}\\
\P$_t$, a=100 &0.57&0.36&0.18&0.06&0.02&0.59&0.42&0.23&0.15&0.10&0.65&0.47&0.40&0.32&0.23&0.72&0.56&0.48&0.28&0.14\\
\P$_c$, a=1 &\textbf{0.62}&0.41&0.23&0.08&0.03&\textbf{0.61}&\textbf{0.44}&\textbf{0.26}&\textbf{0.16}&\textbf{0.11}&0.67&0.49&0.42&0.34&\textbf{0.25}&0.73&0.59&0.51&0.32&0.18\\
\P$_c$, a=10 &0.62&0.43&0.24&0.09&0.04&0.60&0.43&0.25&0.16&0.11&\textbf{0.67}&\textbf{0.50}&\textbf{0.42}&\textbf{0.35}&0.25&\textbf{0.74}&\textbf{0.60}&\textbf{0.51}&0.31&0.17\\
\P$_c$, a=100 &0.61&0.44&0.25&0.09&0.03&0.60&0.43&0.25&0.16&0.11&0.66&0.49&0.42&0.35&0.24&0.73&0.58&0.51&\textbf{0.33}&0.19\\
SOTA&0.62&0.42&0.23&0.09&\textbf{0.04}&0.60&0.44&0.25&0.16&0.11&0.66&0.48&0.41&0.34&0.24&0.73&0.59&0.50&0.31&0.17
\end{tabular}
\caption{Image perturbation robustness of ImageNet models on 288 image size.}
}
\end{table}
\end{landscape}

\section{ImageNet results}

\begin{figure*}[h]
  \centering
  \includegraphics[width=0.95\textwidth]{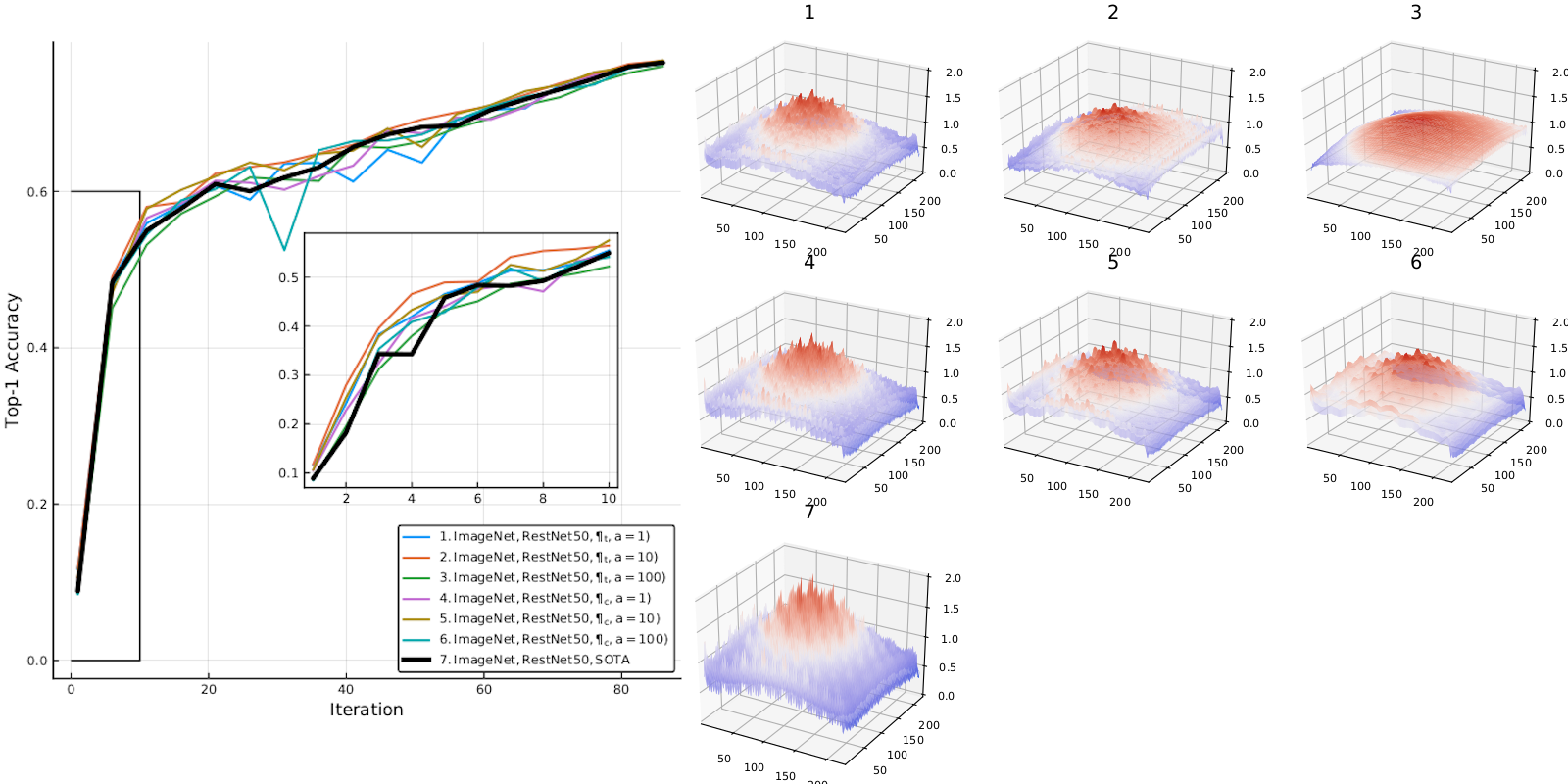}
  \caption{ImageNet with ResNet-50 accuracy and landscape curvature results. The first epochs are zoomed in to show the greater speed of convergence of the \P-regularized objective.}
  \label{fig:imagenet_rc1}
\end{figure*}

\begin{figure}
  \centering
  \includegraphics[width=0.5\textwidth]{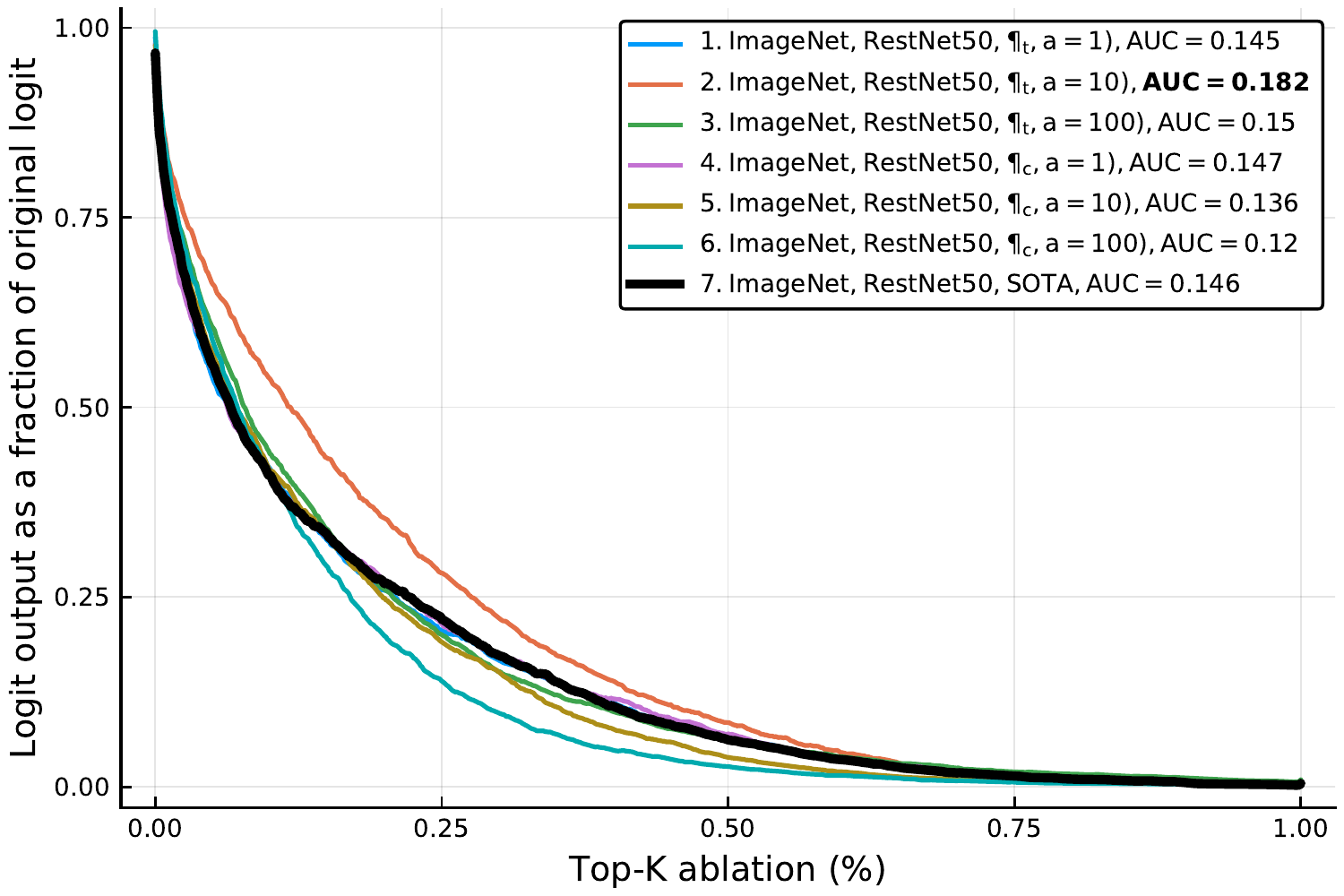}
  \caption{Top-$K$ ablation based on salience map for the different versions of ResNet-50 trained on ImageNet. As shown by the Area Under the Curve (AUC) in the legend, a narrow range of \P$_{\text{t}}$ regularization weight made the model more robust to corruption when compared to SOTA, while too much regularization seemed to have the opposite effect.}
  \label{fig:imagenet_robustness}
\end{figure}

}

\end{document}